\documentclass[conference]{IEEEtran}
\IEEEoverridecommandlockouts
\usepackage{cite}
\usepackage{amsmath,amssymb,amsfonts}
\usepackage{algorithmic}
\usepackage{graphicx}
\usepackage{textcomp}
\usepackage{xcolor}
\usepackage{url}
\usepackage{booktabs}
\def\BibTeX{{\rm B\kern-.05em{\sc i\kern-.025em b}\kern-.08em
    T\kern-.1667em\lower.7ex\hbox{E}\kern-.125emX}}
\usepackage[hidelinks,colorlinks=true,linkcolor=blue,citecolor=blue]{hyperref}

\begin{document}

\title{Answering Students' Questions on Course Forums Using Multiple Chain-of-Thought Reasoning and Finetuning RAG-Enabled LLM}

\author{
\IEEEauthorblockN{Neo Wang$^{1}$ and Sonit Singh$^{1}$}
\IEEEauthorblockA{\textit{$^{1}$School of Computer Science and Engineering, University of New South Wales, Sydney, Australia}\\
\textit{Email: \url{sonit.singh@unsw.edu.au}}
}}

\maketitle

\begin{abstract}
The course forums are increasingly significant and play vital role in facilitating student discussions and answering their questions related to the course. It provides a platform for students to post their questions related to the content and admin issues related to the course. However, there are several challenges due to the increase in the number of students enrolled in the course. The primary challenge is that students' queries cannot be responded immediately and the instructors have to face lots of repetitive questions. To mitigate these issues, we propose a question answering system based on large language model with retrieval augmented generation (RAG) method. This work focuses on designing a question answering system with open source Large Language Model (LLM) and fine-tuning it on the relevant course dataset. To further improve the performance, we use a local knowledge base and applied RAG method to retrieve relevant documents relevant to students' queries, where the local knowledge base contains all the course content. To mitigate the hallucination of LLMs, We also integrate it with multi chain-of-thought reasoning to overcome the challenge of hallucination in LLMs. In this work, we experiment fine-tuned LLM with RAG method on the HotpotQA dataset. The experimental results demonstrate that the fine-tuned LLM with RAG method has a strong performance on question answering task.
\end{abstract}

\section{Introduction}
In large university courses, online student forums (such as Moodle and Ed forum) play a crucial role in facilitating student discussions and resolving academic queries. In the beginning, it is possible for course staff to respond to queries in a timely manner. However, with a high volume of posts, many questions become repetitive, leading to delays in response times and an increased burden on instructors. Students often struggle to find existing answers, and teaching staff must repeatedly address similar concerns. Therefore, a related question recommendation system can help mitigate these challenges by efficiently suggesting previously answered relevant questions, improving student learning experiences, and reducing instructor workload.

In this paper, we developed a question answering system based on large language model (LLM) and retrieval augmented generation (RAG), and also fine-tuned on the relevant course dataset. The system contributes to the development of AI applications in the education domain. By introducing a related question answering system, it promotes a reduction in course staff workload. Teaching staff could focus on more complex or new questions rather than answering repetitive or simple questions. In addition, the developed system could provide an immediate response to student queries anytime (available 24/7) and reduce the need to find similar posts. The proposed system could also improve self-learning in students as they can always find help related to the course whenever they need. Additionally, leveraging LLMs for related question recommendation system could increase search precision, contributing to AI-driven educational support and broader applications such as knowledge-sharing platforms.

This research aims to improve the efficiency of online course forums by developing a question-answering system based on Retrieval-Augmented Generation (RAG), which leverages an external knowledge base to enhance the ability of LLM to generate more accurate and contextually relevant responses. Our system could identify whether a newly posted student question has already been answered in the course forum. If a similar question exists, the system will generate an answer based on the corresponding previous answer. Additionally, multiple chain-of-thoughts reasoning would be implemented to reduce the hallucination of LLM generator.

The specific objectives of this research are to:

\begin{enumerate}
\item To design and implement RAG-based system which could help the LLM generator produce a better answer
\item To incorporate multiple chain-of-thought (COT) reasoning to improve the performance of student question answering system
\end{enumerate}

This paper is structured as follows: The relevant background and the related work of question answering system are introduced in section ~\ref{sec:rl}. In section~\ref{sec:methodology}, we describe our approach to mitigate the current limitation. The details of experiment are provided in section\ref{sec:results} with results analysis and discussion in section\ref{sec:discussion}. Finally, in section \ref{sec:conclusion}, we summarise the benefits and limitations of our proposed question answering system, including potential future directions that could further improve this work.

\section{Background and Related Work}~\label{sec:rl}

In this section, we lay the foundation of building blocks needed to understand the literature and also based on which our proposed methodology is based on. We also provide an overview of relevant work related to question answering system.

\subsection{Recent advancements in Natural Language Processing}

Natural Language Processing (NLP) is a sub-field of Artificial Intelligence (AI) that aims to develop computational algorithms or tools to analyse or synthesise natural language and speech. NLP spans two main tasks, namely, natural language understanding (NLU) and natural language generation (NLG). In general, the development of NLP can be divided into three generations. Early methods were rule-based, which are mostly based on rules made by humans to finish several tasks\cite{b14}. For example, if the input contains 'Hello', the rules told model to respond with 'Hi'. This method could perform well on specific tasks if the rules are designed well. However, ruled-based methods require high accuracy of rules made manually and it would be expensive. Additionally, even if a model with well-designed rules performs performs good on a specific task, it is difficult to generalise well on other similar tasks. This means we need to design rules from scratch for a new tasks. To address the generalisation issue of rule-based model, the statistical models were introduced, such as N-gram\cite{b15}. In the stage, features could be utilised to achieve different tasks.

A key task in NLP is to convert human natural language into a representation that can be process by computer. One hot vector\cite{b16}, for example, was a popular method to represent words. It is a simple but effective method to store a word in computer. But, this method have limitations in terms of high dimensionality and also do not capture context. Every new word means we need to create a new one hot vector for it, and the size of each vector equals to the size of vocabulary. In addition, one hot vector cannot extract specific features of entities.

Word2Vec\cite{b17} utilises features to convert word to vector, and the size of vector just depends on the number of features rather than the whole vocabulary. Word2Vec learns dense representations of words, one for centre word, another for context words. Skip-gram is using centre word to predict the context word within finite context window and continuous bag of word (CBOW) is using context word to predict the centre word. GloVe\cite{b18} constructs the vector based on co-occurrence. It aims to reduce the dimensionality of word representation and improved the efficiency of NLP models.

With the development of deep learning, several neural networks, such as recurrent neural network (RNN) and Long Short-term Memory (LSTM), perform well on many NLP tasks. These models promote the application of neural network in NLP domain. But the performance is still limited by the capability of models. Therefor, self-attention mechanism is introduced.

Transformer model\cite{b7} proposed the idea of self-attention. This mechanism significantly improves results on many NLP tasks. It simulates the human attention in the structure to leverage the previous networks. The
Bidirectional Encoder Representations from Transformers (BERT) model\cite{b20} proposed a novel structure based on the pre-training of transformer. Concretely, the text representations are pre-trained bi-directionally rather than single directional. This method creates various state-of-the-art models on numerous NLP tasks including Question-answering, machine-translation. 

Building upon the Transformer architecture\cite{b7}, recent years have witnessed significant advancements in NLP, driven by the emergence of Large Language Models (LLMs)\cite{b6}. These models, often consisting of billions of parameters, are trained on massive corpora and exhibit impressive capabilities in tasks such as question answering, summarisation, and language generation. Notably, LLMs like OpenAI’s GPT series are based on a decoder-only Transformer architecture\cite{b19}.

\begin{figure}[htbp]
\centering{\includegraphics[scale=0.2]{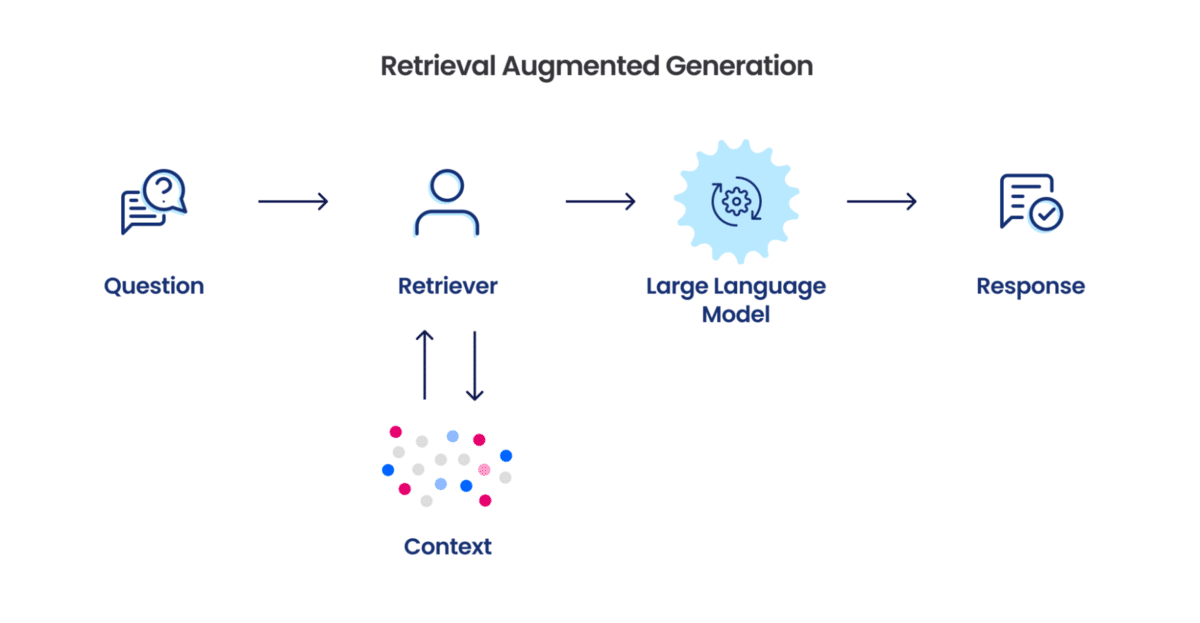}}
\caption{Structure of RAG system}
\label{fig:RAG_structure}
\end{figure}

Despite their powerful generative abilities, LLMs are limited by the fixed knowledge encoded during pre-training. This has led to the development of hybrid frameworks that combine LLMs with external knowledge sources to enhance their performance and factual accuracy. One prominent approach is Retrieval-Augmented Generation (RAG)\cite{b8}, which integrates information retrieval mechanisms into the generation process to dynamically access relevant documents at inference time. Retrieval-Augmented Generation (RAG) introduces an external retrieval mechanism to LLMs. This strategy aids LLMs to avoid hallucination and improve the performance of LLMs on generating precise content\cite{b8}. RAG systems consist of three main components: retriever, generator and augmentation methods. With the release of LLMs such as ChatGPT, RAG systems can be leveraged with LLMs to improve its results, and RAG system performs well on reducing hallucination of LLMs. On other hand, the LLMs with RAG could easily be updated without re-train from scratch. This means that it could be economical for its applications. Moreover, the structure of RAG indicates that it can be easily deployed locally. So there would be no concern about privacy. Figure~\ref{fig:RAG_structure} shows the RAG structure. 

Recently, supervised fine-tuning (SFT)\cite{b26} is a method that adapt pre-train models on specific tasks through training them on labelled dataset.  It enables the model to align with task-specific objectives, such as classification or summarisation, while retaining general language understanding. During the SFT process, we can fine-tune all parameters or partial parameters, offering a trade-off between performance and computational efficiency.

\subsection{Related Work}
Several studies explored the application of LLMs in education domain\cite{b5}. These studies focus on the Chatbot in supporting students self-learning, yet few evaluated their methods on well-designed metrics. In particular, these papers only validate their result on a few samples. RAGBench addresses this problem by proposing the TRACe evaluation framework\cite{b1}. The study combines twelve different datasets to evaluate RAG systems and also measures Utilisation, Relevance, Adherence, and Completeness to analyse system performance. Although the study only evaluated GTPs models which is less economical than open-source model, it provided a novel framework for evaluation the performance of RAG system.

In RetLLM\cite{b10}, authors proposed a retrieval-augmented LLM for question answering on students discussion forums. It demonstrated the effectiveness of the appropriate prompt on Question-Answering system of education domain. Moreover, prompts are designed to require that LLMs do not directly generate solutions, such as code or answer for assignment. Instead, RetLLM-E would output hints and suitable guides for students to solve their questions. The authors also combine human evaluation and ROUGE \& BERTScore to evaluate the model in multiple dimensions. Also, authors compared LLM with retrieval and LLM without retrieval to justify the advantages of RAG method. Preconditions are applied in RetLLM-E to mitigate hallucination. However, RetLLM-E is still limited by the quality of data source. Concretely, it might produce wrong response when there is not related documents retrieved. In addition, it performs differently on different type of questions. Although, the paper proposed a student-oriented question answering system through producing hints rather than direct answers for students for some specific question types, it lacks method to automatically evaluate the quality of hints.

In another study\cite{b2}, authors proposed RAG model on task-specific domain through combining pre-trained seq2seq model and pre-trained retriever. The RAG model outperforms state-of-the-art models on open domain question-answering tasks and approaches similar performance on other tasks such as question generation and fact verification. Moreover, it avoided several unnecessary costs including training from scratch on a new dataset, dependency on specific documents for extractive or abstractive tasks. The proposed model not only utilises the documents retrieved to improve the reliability, but also generates reliable responses with parametric memory when the corresponding resources are insufficient. The paper provides a retrieval augmented method to mitigate the hallucination. It not only highlights the reduction on hallucination, but also avoid the cost on re-training model when the dataset have to be updated. However, the study has limitations in terms of the quality of documents as inferior resources may cause bias, misleading or abuse content.

Li \emph{et al.}\cite{b9} found the main challenges on deploying AI on education domain are hallucination and difficulties on updating LLMs. Thus, they introduce RAG method to improve this situation. Authors identified the high API cost if using paid LLMs such as ChatGPT and highlighted the need to explore more affordable options for applying LLMs in the education domain.  Sharma \emph{et al.}\cite{b4} proposed a novel RAG framework for domain-specific question answering task. They utilised users' activities, such as log clicks to improve the performance of retriever. Also, they define a relevancy metric to rank the query-document during training retriever model. In terms of generator, LLM with few-shot prompt are introduced to their framework. Authors use not only the grounded sources to fine-tune the LLM, but also add some negative documents (moderately dis-similar with others) to strengthen the recall and robustness of generator. In addition, query augmentation are implemented in the framework to mitigate the ambiguous expression. Last but not least, they utilise a Named Entity Removal Module to ensure the privacy of users. The proposed method provided improvement on LLM generator to augment the question answering system, potentially mitigating the hallucination.

In\cite{b11}, the authors proposed an AI Discussion Assistant (AIDA) system to support instructors' work and analysed the significance of AIDA in education domain. The study addresses the fact that AI tools cannot directly replace the real instructors in education domain. Thus, they use AIDA to support instructors in answers queries of student in the online forum. The AIDA offers several options including directly generating answer, retrieve previous related materials, generating response with context retrieved. The instructors could select the appropriate work mode to generate final answer. In addition, they implemented an instructor-in-the-loop method to help LLM generate a better answer. Concretely, there might be some deletions or additions based on the generation of LLM. This strategy significantly improves the quality of final answers. Also, the system is evaluated via the number of modifications, with lower modifications result in greater accuracy. However, there are still limitations of the proposed AIDA system as it does not support multimodal inputs and the relevant context retrieved is manually operated by instructor, which could be automated. 

Boros \emph{et al.}\cite{b12} proposed a novel recipe, \emph{Sherlock}, for question-answering task with affordable LLM. The authors firstly select several open-source LLMs as their base model including instruct model and non-instruct model. Then they perform supervised fine-tuning on two different datasets, and this yields two results. The two models from the previous step are merged with a new different model. Then a Direct Preference Optimisation is implemented on both of them. Finally, they developed two different model: one with RAG and another without RAG. Those strategies notably increase the performance of Sherlock. In addition, the final step (ablation test) demonstrates the advantages of RAG method. Sherlock achieves a great result on test set with limited parameter size. This means Sherlock successfully saves expenditure on LLM API and computational resources. However, there are limitations in their study, including model can't correct n-grams for RAG part and experiments on specific base models.

Miladi \emph{et al.}\cite{b13} focused on evaluating the effectiveness of LLM with RAG. The authors compared control group (CG) without support from AI tools and experimental group (EG) with support from AI tools. Concretely, the evaluation was finished via pre-test, post-test and System Usability Scale (SUS) questionnaire. The result demonstrated the ability of AI tools on supporting students on acquiring new knowledge and the usability of a conversational agent. However, the experiment is limited due to the small group size. Additionally, there is no ablation test to prove the effect of the RAG method. Yoran \emph{et al.}\cite{b21} proposed a novel method - MultiChain Reasoning (MCR) which could meta-reason through multiple chains of thoughts. Firstly, the authors demonstrated the current limitations of chain-of-thought reasoning method on question answering. The result shows though the single chain-of-thought could provide useful information, it might generate wrong answer due to the incorrect direction of thoughts. In addition, this approach would take advantage of the evidence from the multiple chains to generate the final answer instead of directly aggregating them together. The authors mainly experiment this method with implicit reasoning and explicit reasoning. It turns out that their model outperforms all related previous models. Additionally, the final result is improved through combining their model with another model - self-consistency (SC)\cite{b22} from others' study. The study did not apply any supervised fine-tuning method to improve the performance of the model. Also, the evaluation of the explanation quality of their model was done manually which would be inefficient and costly.

\section{Methodology}~\label{sec:methodology}
According to the literature review in the previous section, we found that various studies have explored the integration of LLM into education. However, the majority of them are simply adopting LLM API to build a question answering system. Although, this approach takes advantage of the pre-trained knowledge of the LLM, it does not take into account the specific course related knowledge, which is highly valuable and unique to particular course in a particular university. Moreover, using paid LLM APIs such as ChatGPT could be very expensive when thousands of students were using the forum and every call to the API costs. Therefore, it is important to explore alternatives, such as open-source LLMs to apply them in the education settings. Also, we observe that most of the existing work in literature suffer from hallucination problem. Finally, existing work on Question-Answering systems in education settings focused on providing the final answer, which can sometimes be deteriorating from students' learning perspective. To overcome all the above highlighted limitations, we propose a Question-Answering system based on open-source LLM and making use of local course content using RAG method and fine-tuning it so that the system can learn content specific to the course content. Moreover, to overcome the limitation of not giving directly the final answer, we integrated multiple chain-of-thought reasoning approach. The proposed system not only is cost-effective, can provide step by step response to students' question, but also overcomes the problem of hallucination by enhancing the capability of the LLM generator.

\begin{figure*}[htbp]
    \centering{\includegraphics[scale=0.6]{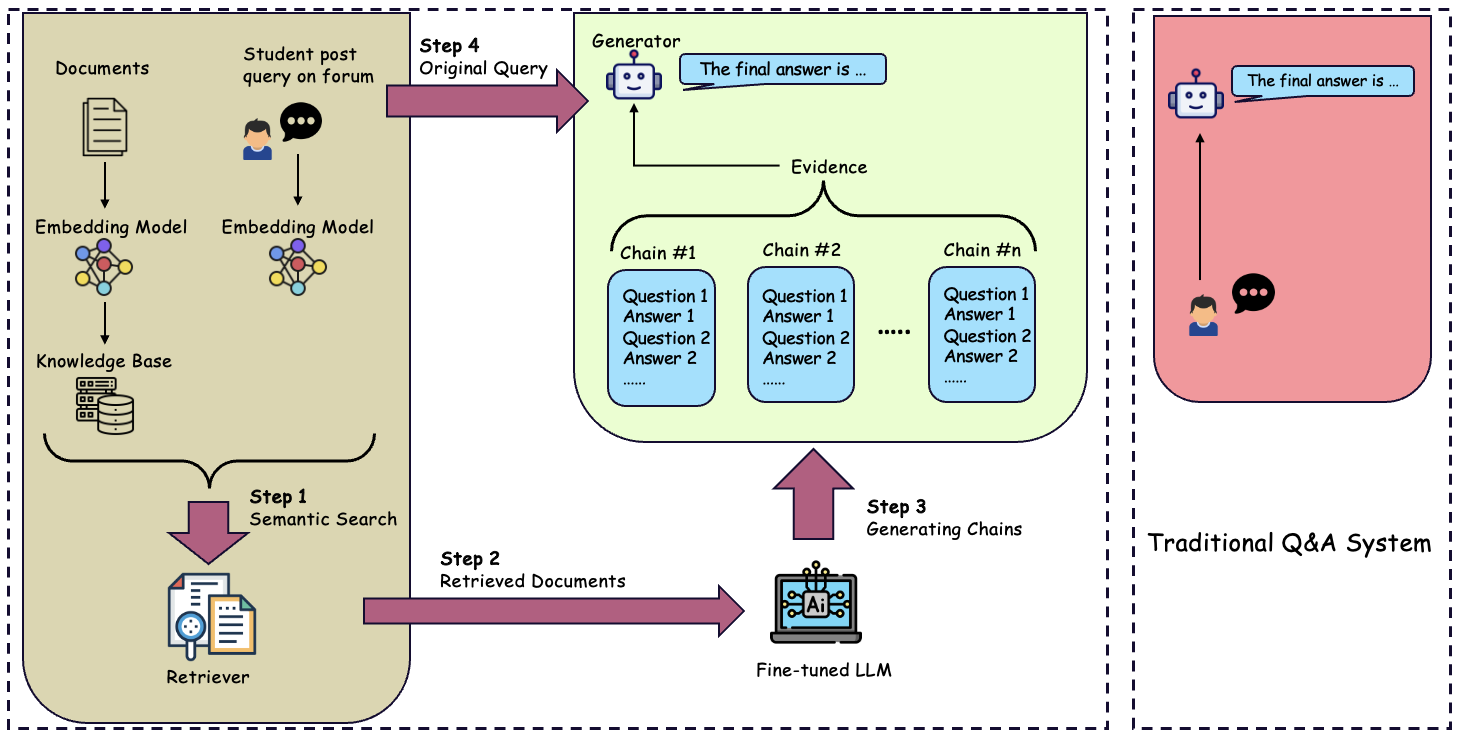}}
    \caption{Multi-chains Reasoning with RAG VS Traditional Question-answering System}
    \label{fig:frame_figure}
\end{figure*}

\begin{figure*}[htbp]
    \centering
    \includegraphics[scale=0.9]{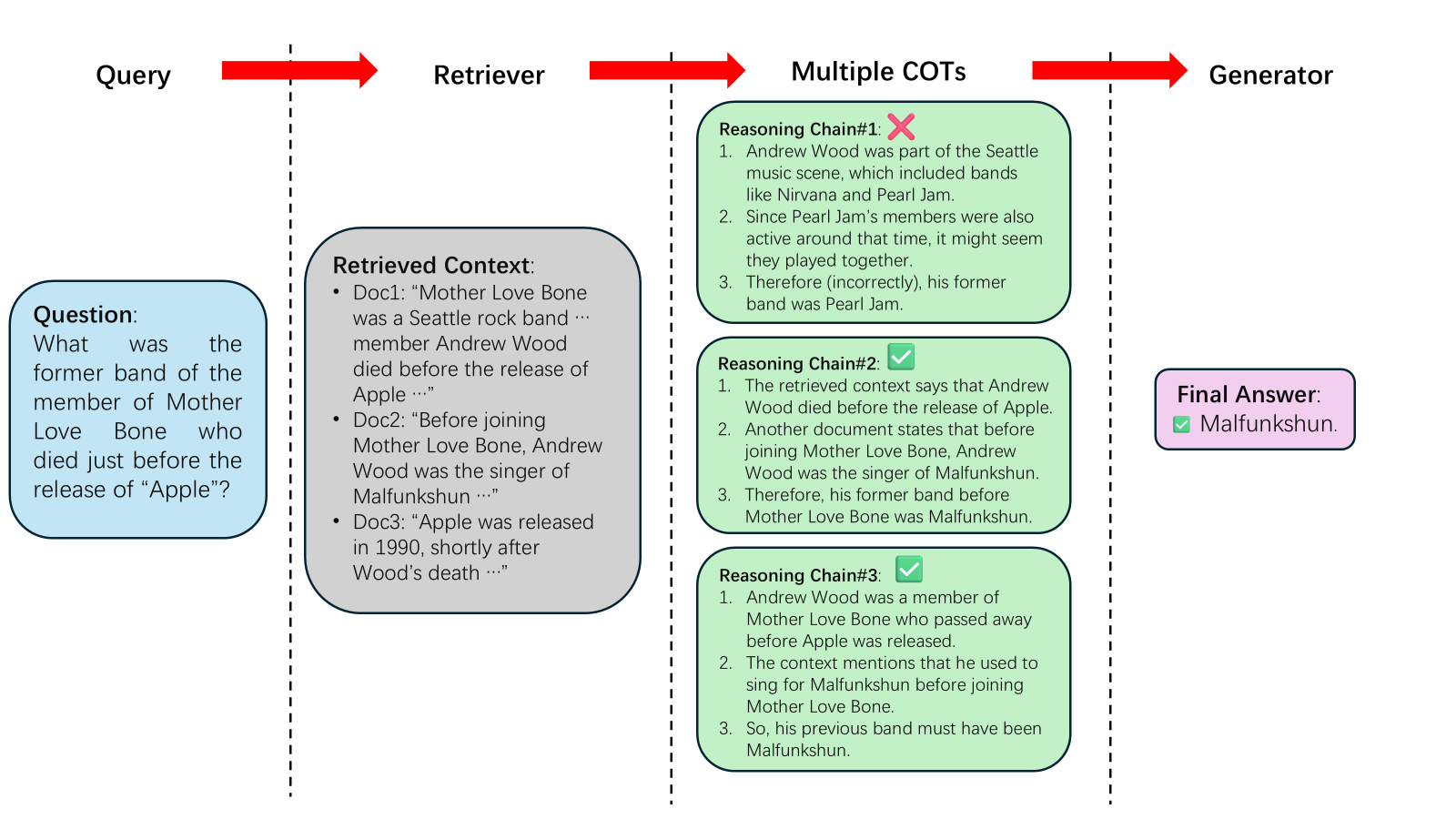}
    \caption{Example of Multi-chains Reasoning.}
    \label{example_mcr}
\end{figure*}

Our proposed finetuned RAG-enabled LLM using multiple chain-of-thought reasoning architecture for answering students' questions on course forums is shown in Figure~\ref{fig:frame_figure}. We propose a method to answer students' questions on course forum based on the past offerings of the course data. Our question answering system constructs a local knowledge base for course based on its current content and past questions and answers. When students post their questions on forum, the system would search the top-k similar documents or questions based on the input queries. Then the retrieved documents will be passed into the first LLM to generate multi-chains. The results of multi-chains are combined with the original questions and the designed prompt. The results of the multi chain-of-thought, question and the prompt, are given as the input to the second LLM (generator) and the generator will leveraging the evidence to the final answer. In summary, first the LLM is prompted to produced multiple chain-of-thought reasoning intermediate processes based on retrieved documents and the original question. After this, the reasoning steps are collected and passed to the LLM. Finally, after receiving all intermediate processes, the LLM would utilise them as evidence to generate the final answer. We illustrate the effectiveness of the multiple chain-of-thought (COT) reasoning and the entire process in Figure~\ref{example_mcr}. The question (see Figure~\ref{example_mcr}) seeks the information about which band a member of Mother Love Bone belonged to before his death, just prior to the release of ``Apple”. To solve this query, three documents are returned by the retriever. The first document told the information about band member, the second told the previous band of this member, and the third document told the data of the release of ``Apple”. Then the first LLM produced three chains to infer the final answer. The first COT generated an incorrect answer, but the remaining two both generated the correct result. Finally, the second LLM generator produced the correct answer based on the previous results. From this simplified example, we could observe that only one chain-of-thought reasoning might lead to a wrong result. But multiple chain-of-thought reasoning could avoid this situation as much as possible. In the following sections, we provide more details about various blocks of our proposed architecture. 

\subsection{Retriever}
Each document in our dataset is converted into embedding vector space. Similarly, the student questions are also expressed as vector representations. Then the retriever would search the top-k documents which have similar semantics to input queries through search function provided by FAISS\cite{b23} library. Then the retriever would pass the retrieved documents to generator as context.

\subsection{Generator}
We adopted Llama-3.2-3B-Instruct\cite{b24} as our base model due to its excellent performance on several tasks.

\paragraph{Zero-Shot LLM generator with RAG}
We design a simple prompt to ask LLM generate answer according to the retrieved documents. In our experiments, we found that a long complex prompt might negatively impact the LLM to generate the correct answer. The retrieved documents are too long for LLM to understand the whole context. Therefore, the prompt is designed simply to reduce the complexity of context.

\paragraph{Finetuned LLM generator with RAG}
To improve the performance, we implemented LoRA method to finetune the LLM generator. The loss of prompt parts is set as a small negative constant, so the model could only learn the response part.

\subsection{Embedding model}
We used \textit{all-MiniLM-L6-V2} as an \emph{embedding} model. It is a sentence embedding transformer-based model, and can map input sentences 384 dimensional dense vector space. The embedding model is implemented to semantically search the related documents.

\subsection{Dataset}

The HotpotQA dataset mainly consists of questions, answers and a set of documents for retrieval. The HotpotQA dataset\cite{b3} is a large-scale, comprehensive dataset designed for multi-hop question answering that requires complex reasoning over multiple retrieved documents. The documents used for the retrieval component in our system are derived from Wikipedia articles, consistent with the original dataset setup.

According to the analysis of previous study\cite{b3}, there are more than 20 different types of questions in the HotpotQA dataset. To simplify the question classes, we use the corresponding answer types to redefine the question classes. The distribution of various question types in the HotpotQA dataset is given in Table~\ref{tab:question_in_dataset}. The distribution of samples across the training, validation, and test subsets of the HotpotQA dataset is summarised in Table~\ref{tab:HotpotQA_Stats}.

\begin{table}[]
    \centering
    \caption{Questions in HotpotQA}
    \begin{tabular}{lc}
        \toprule  
        Question Type & Percentage\\
        \midrule  
        Person & 30\%\\
        Group / Org & 13\%\\
        Location & 10\%\\ 
        Date & 9\%\\ 
        Number & 8\%\\
        Artwork & 8\%\\
        Yes/No & 6\%\\
        Adjective & 4\%\\
        Event & 1\%\\
        Other proper noun & 6\%\\
        Common noun & 5\%\\
        \bottomrule 
    \end{tabular}
    \label{tab:question_in_dataset}
\end{table}

\begin{table}[]
    \centering
    \caption{HotpotQA Dataset Statistics}
    \begin{tabular}{lc}
        \toprule
        Subset& \#Samples\\
        \midrule
        Train& 1,880\\
        Valid& 424\\
        Test& 390\\
        \bottomrule
        \end{tabular}
     \label{tab:HotpotQA_Stats}
\end{table}

\subsection{Experimental Setting}
We implemented our proposed architecture in Python programming language using PyTorch deep learning framework. Experiments were conducted on a GPU cluster to accelerate computing. Various hyperparameters settings are summarised in Table~\ref{tab:lora_para}. There were a total of 1,805,760,512 parameters, with 2,293,760 trainable parameters. By implementing LoRA method, we only train the 0.127\% parameters to improve the efficiency during fine-tuning. 

 \begin{table}[]
    \centering
    \caption{Parameter setting}
    \begin{tabular}{llll}
        \toprule  
        Parameter& Value\\
        \midrule  
        LoRA rank & 8\\
        LoRA alpha & 16\\
        LoRA dropout & 0.05\\ 
        Learning rate & 2e-9\\ 
        Training epochs & 3\\
        Training batch size & 2\\
        Evaluation batch size & 2\\
        Gradient accumulation steps & 4\\
        \bottomrule 
    \end{tabular}
    \label{tab:lora_para}
\end{table}

\subsection{Evaluation Metrics}
Following previous studies on question-answering systems, F1 score is primarily used to evaluate our model on this dataset, complemented by BLEU for fluency/overlap and Semantic Similarity for meaning preservation.

\paragraph{F1-score}
F1 score is a metric used to evaluate the overall accuracy of a model by considering both precision ($P$) and recall ($R$) through their harmonic mean. In our experiments, we use the macro-average F1 score to assess our question-answering system's performance across all instances.

The formula for the F1 score is:
\begin{equation}
F1 = 2 \cdot \frac{P \cdot R}{P + R},
\label{eq:F1}
\end{equation}
Where Precision and Recall are defined as:
\begin{equation}
P = \frac{\text{True Positives}}{\text{True Positives} + \text{False Positives}},
\label{eq:Precision}
\end{equation}
\begin{equation}
R = \frac{\text{True Positives}}{\text{True Positives} + \text{False Negatives}}.
\label{eq:Recall}
\end{equation}
The macro-average F1 score is calculated by first computing the F1-score for each QA pair and then taking the average across all pairs.

\paragraph{BLEU}
Bilingual Evaluation Understudy (BLEU) score is an algorithm primarily designed for machine translation (MT) quality assessment but is frequently adopted for text generation tasks like Question Answering to measure the similarity between the model's generated text and a set of reference texts.

BLEU relies on calculating the proportion of matching $n$-grams (typically up to $N=4$) between the candidate and reference texts, applying a Brevity Penalty (BP) to penalise overly short outputs.

The overall BLEU score is calculated as:
\begin{equation}
\text{BLEU} = \text{BP} \cdot \exp \left( \sum_{n=1}^{N} w_n \log P_n \right),
\label{eq:BLEU}
\end{equation}
Where $P_n$ is the modified $n$-gram precision, $w_n$ are the weights for each $n$-gram order, and BP is the Brevity Penalty, which severely penalises candidates shorter than the reference texts:
\begin{equation}
\text{BP} =
\begin{cases}
1 & \text{if } c > r \\
\exp\left(1 - r / c\right) & \text{if } c \le r
\end{cases},
\label{eq:BP}
\end{equation}
where $c$ is the length of the candidate sentence and $r$ is the effective reference length that is closest to $c$.

\paragraph{Semantic Similarity}
Semantic Similarity evaluates how close two text segments (the model's generated answer and the ground-truth answer) are in meaning, rather than just word overlap. This metric is crucial for capturing paraphrases and meaning equivalence that lexical metrics might miss.

Semantic Similarity is typically computed using Sentence Embeddings derived from powerful Pre-trained Language Models (PLMs) such as Sentence-BERT (SBERT). The process involves generating embedding vectors ($\mathbf{A}$ and $\mathbf{B}$) for the answers and calculating the Cosine Similarity between them.

The formula for Cosine Similarity between two $n$-dimensional vectors $\mathbf{A}$ and $\mathbf{B}$ is:
\begin{equation}
\text{sim}(\mathbf{A}, \mathbf{B}) = \frac{\mathbf{A} \cdot \mathbf{B}}{\|\mathbf{A}\| \|\mathbf{B}\|} = \frac{\sum_{i=1}^{n} A_i B_i}{\sqrt{\sum_{i=1}^{n} A_i^2} \sqrt{\sum_{i=1}^{n} B_i^2}}.
\label{eq:Cosine}
\end{equation}

A higher cosine similarity value (closer to $1$) indicates greater semantic alignment between the generated and reference answers.

\section{Results}\label{sec:results}

In this section, we provide experimental results based on the evaluation metrics on the HotpotQA dataset. 

\subsection{Retrieval Augmented Generation}
To highlight the effectiveness of RAG method, an ablation study was conducted. The results reported in the original study on HotpotQA dataset\cite{b3} was considered as a baseline. The experimented results, as shown in Table~\ref{tab:ragresult} indicate that there is a significant increase on F1 score for Llama-3.2-3B-Instruct with RAG method. We further extended our experiments by finetuning the Llama-3.2-3B-Instruct model with and without RAG. The results as given in  Table~\ref{tab:ragresult} demonstrate significant improvement in a finetuned model with RAG. 

\begin{table}[]
    \centering
    \caption{Result of ablation study for RAG approach}
    \begin{tabular}{llll}
        \toprule  
        Model& F1\\
        \midrule
        Benchmark of HotpotQA\cite{b3}& 34.4\\
        Llama-3.2-3B-Instruct without RAG& 19.0\\
        Llama-3.2-3B-Instruct with RAG & 26.8\\
        Finetuned Llama-3.2-3B-Instruct without RAG & 59.6\\
        Finetuned Llama-3.2-3B-Instruct with RAG & \textbf{62.2} \\ 
        \bottomrule 
    \end{tabular}
    \label{tab:ragresult}
\end{table}

\subsection{Multiple Chain-of-thought reasoning}
In terms of multiple chain-of-thought reasoning, we compared the performance of multiple chain-of-thoughts reasoning for different chains. 

\begin{table}[h]
    \centering
    \caption{Result of Multiple Chain-of-thoughts Reasoning}
    \begin{tabular}{lccc}
        \toprule
         Models& F1& BLEU& Semantic Similarity\\
        \midrule 
        MCR(1 chain)& 30.7& 12.0& 64.6\\
        MCR(2 chain)& \textbf{32.0} & \textbf{12.8} & \textbf{66.6} \\
        MCR(3 chain)& 28.5& 10.7& 66.3\\
        MCR(4 chain) & 27.5& 10.8& 65.4\\
        \bottomrule 
    \end{tabular}
    \label{tab:base_mcr}
\end{table}

To align with the previous work, we use the same retriever (FAISS) and generator (Llama-3.2-3B-Instruct). When the number of chains is set as 2, our baseline model has achieved the best performance. Compared with the result of Llama-3.2-3B-Instruct with RAG, we can observe a significant increase in F1 score. This demonstrates that the multiple chain-of-thought reasoning enhance our system to generate a better answer to question.

\section{Discussion}\label{sec:discussion}
Large Language Models (LLMs) have the potential to transform how we support students in their learning in a university settings. Our results demonstrated that fine-tuning RAG-enabled LLM can learn course specific content and can answer students' questions on course forums more effectively compared to without fine-tuning and without using any course-related database. Furthermore, our experiments demonstrated the need to have multiple chain-of-thought reasoning so that the developed system can answer students' queries by giving hints in multiple turns, without directly providing answer, which will support students' learning.  

As given the results section, we found that there is a significant improvement in results when we use course content and previous questions accompanied with their answers from past course offerings as a local knowledge base. Having a local knowledge base not only provide more relevant content but also provide more control in terms of answers given to students. Although pretrained LLMs can respond to students' queries but we found that the answers given by LLM to students' questions are very generic and do not reflect the depth and scope of the course content taught at a particular institution. Using the RAG approach for retrieving relevant document from a local knowledge base helps in providing improved answers. As given in results Table~\ref{tab:ragresult}, we can see a significant increase on F1 score for Llama-3.2-3B-Instruct with RAG method. Second, although existing studies showed that LLMs can answer students' questions but we found that providing direct answer is not good from students' learning perspective. It would be more effective to provide step-by-step reasoning so that students can understand how we reach to the final answer. Moreover, it is important to link various evidences to reach out to the final answer. According to the results of multi chain-of-thought reasoning, (see Table~\ref{tab:base_mcr}), we observed that it is not always beneficial to simply increase the number of chains. We speculate that the reason why the model perform best when the number of chains is 2 is more chains might produce more incorrect content. We observe that single chain might produce the incorrect answer. Hence, it is of importance to decide the number of chains. If we increase the number of chains, there would be a drop in their overall performance. Consequently, directly increasing the quantity is not always useful. Otherwise, this not only increases the cost of multiple COT reasoning, but also might decrease the overall performance. Therefore, deciding an appropriate setting for the number of chains is crucial to improve the multiple chain-of-thought reasoning on question-answering system. Based on our ablation study, we found that having two chains provide best results, demonstrating that more that two chains significantly increase the complexity of reasoning. 

\section{Conclusion}\label{sec:conclusion}
In this work, we proposed a novel architecture combining finetuned LLM with RAG knowledge and chain-of-thought reasoning to answer students' questions on the discussion forums. The experiments on a public question answering dataset, HotPotQA, showed that RAG knowledge, when combined with finetuned LLM, is effective in providing better answering to students' question on course discussion forums. The multiple chain-of-thought reasoning also enhances the quality of generated responses when the number of chains is correctly set.

In our future work, we aim to further improve the proposed system by focusing on multimodal course content as we found that most course content is not just text-based. Thus, multimodal retriever would be considered to improve our question answering system. Furthermore, we could improve the effectiveness of retriever as the correctness of the final answer is highly dependent on the quality of retrieved documents from the local knowledge base using the RAG approach. Finally, there are many novel methods such as Reinforcement Learning with Human Feedback (RHLF) that have potential to strengthen our proposed framework by bringing human-in-the-loop. We believe that having automated students' question answering system for courses with human oversight have huge potential to support students' learning and taking away workload from instructors. 

\section*{Acknowledgement}
This research was supported by Katana, the high performance computing facility at the University of New South Wales. The authors also acknowledge the financial support provided by the School of Computer Science and Engineering for API and cloud services used in the development of the question answering system. 


\end{document}